# Title: Deep Learning-Based Detection of the Acute Respiratory Distress Syndrome: What Are the Models Learning?

Gregory B. Rehm MS
Department of Computer Science
University of California, Davis
grehm@ucdavis.edu

Chao Wang Ph.D
Department of Computer and Electrical Engineering,
University of California, Davis
chaowang.hk@gmail.com

Irene Cortes-Puch MD
Department of Pulmonary, Critical Care, and Sleep Medicine
University of California, Davis
icortespuch@ucdavis.edu

Chen-Nee Chuah, Ph.D
Department of Computer and Electrical Engineering,
University of California, Davis
chuah@ucdavis.edu

Jason Adams, MD MS (corresponding author)
Department of Pulmonary, Critical Care, and Sleep Medicine
University of California, Davis
jyadams@ucdavis.edu

Word count:

# Key Points:

**Question:** Does deep learning help improve the performance of traditional machine learning ARDS detection models, and if so, what features does it utilize to do so?

**Findings:** In this retrospective study, we find deep learning improves upon traditional machine learning for ARDS detection by utilizing similar information as traditional machine learning models and difficult to manually featurize high-frequency waveform data.

**Meaning:** Our results highlight the utility of deep learning at improving upon featurization of physiologic waveform data for disease detection, and highlights interpretability methods to understand what deep learning models are learning with physiologic waveform data.

**Structured Abstract: (350 word limit)**

**Importance:** The acute respiratory distress syndrome (ARDS) is a severe form of hypoxemic respiratory failure with in-hospital mortality of 35-46%. High mortality is thought to be related in part to challenges in making a prompt diagnosis, which may in turn delay implementation of evidence-based therapies. A deep neural network (DNN) algorithm utilizing unbiased ventilator waveform data (VWD) may help to improve screening for ARDS.

**Objective:** To (i) develop a deep learning (DL) model that enables ARDS screening using raw VWD and improves the accuracy of traditional machine learning (ML) models and (ii) to examine the interpretability of the deep learning model.

**Design:** Retrospective, observational cohort study. Subjects were selected from a prospective cohort study enrolling subjects between 2015 and 2020. Subjects were chosen for this study based on pathophysiology and existence of at least 1hr of VWD after the first 24 hours of intubation or the first 24 hours after ARDS diagnosis.

**Setting:** Academic medical center intensive care units (ICU).

**Participants:** Adults admitted to the ICU requiring invasive mechanical ventilation (MV), including 50 patients with ARDS and 50 patients with other indications for MV other than ARDS.

**Exposures:** A deep learning algorithm.

**Main Outcomes and Measures:** The primary measurement of model performance was area under the curve (AUC) of the receiver operating characteristic (ROC) curve. Measures of accuracy, sensitivity, specificity, positive predictive value (PPV), and negative predictive value (NPV) are also reported.

**Results:** We first show that a convolutional neural network-based ARDS detection model can outperform prior work with random forest models in AUC ($0.95\pm0.019$ vs. $0.88\pm0.064$), accuracy ($0.84\pm0.026$ vs $0.80\pm0.078$), and specificity ($0.81\pm0.06$ vs $0.71\pm0.089$). Frequency ablation studies imply that our model can learn features from low frequency domains typically used for expert feature engineering, and high-frequency information that may be difficult to manually featurize.

**Conclusion and Relevance:** Deep learning creates more accurate models compared to traditional machine learning when creating ARDS screening models using VWD. Further experiments suggest that subtle, high-frequency components of physiologic signals may explain the superior performance of DL models over traditional ML when using physiologic waveform data. Our observations may enable improved interpretability of DL-based physiologic models and may improve the understanding of how high-frequency information in physiologic data impacts the performance our DL model.

# Introduction

The acute respiratory distress syndrome (ARDS) is a subtype of hypoxemic respiratory failure that affects over 10% of all patients in the intensive care unit (ICU) and 23.4% of all patients receiving mechanical ventilation.[1] Outcomes for patients with ARDS are poor with hospital mortality rates ranging from 34.9-46.1% depending on disease severity.[1] Despite its prevalence, ARDS has proven difficult to diagnose[2] and ARDS patients are characterized by relatively low use of proven life-saving therapies,[3–5] which may contribute to the high mortality rates observed for ARDS.[1,6–9]

To improve ARDS identification, multiple studies have researched automated screening methods using electronic health record (EHR) and imaging data.[10–19] However, these methods displayed a number of limitations, including dependence on provider ordering and documentation practices, that may limit their generalizability for wider deployment.[2,10] To address these issues, prior work has investigated the feasibility of using ventilator waveform data (VWD) to screen for ARDS.[20] VWD consists of highly sampled (50 Hz) air flow and pressure time series data recorded by mechanical ventilators and is advantageous because it is ubiquitous in all patients receiving MV. VWD is also continuously sampled making it amenable to real time analytics, which contrasts with many EHR data types that are sampled infrequently or must be manually updated by provider inputs. Furthermore, VWD is unbiased by human subjectivity[2,10,21–23] because it consists of machine sampled pressure and flow observations. Finally, VWD is rich with embedded physiologic information such as respiratory system compliance and airway resistance that may be useful for automated detection of acute causes of respiratory failure including ARDS.[24–26]

A potential limitation of prior work using VWD for ARDS detection is that it used traditional machine learning algorithms like random forests (RF) with expert-informed waveform featurization.[20] Expert-informed feature engineering is common in traditional machine learning and may aid in understanding model behavior but the use of human-interpretable features may bias models in unanticipated ways and may potentially constrain performance.[20,27] To solve these limitations, researchers have applied deep learning (DL) approaches to automatically learn high-level features from the input data.[28] This approach has been used widely in image processing and in other time series analytics work.[29–34] In this study, we hypothesized that DL models would improve upon traditional ML models in discriminating between patients with ARDS and non-ARDS indications for MV. To investigate our hypothesis, we first developed DL models capable of screening for ARDS using only VWD. We analyzed performance of two commonly used DL networks used for this task, convolutional neural networks (CNN) and long short-term memory (LSTM) networks and compared DL performance to that of a previously published traditional ML models. Finally, we investigated the mechanisms by which our best-performing DL model was able to learn features from raw VWD that contributed to the differential performance observed between DL and traditional ML models.

# Methods

Cohort

All data for this study were collected as part of an IRB-approved, observational study in the ICUs at UC Davis Health, an academic tertiary care medical center in Northern California. All patients or surrogates provided informed consent for data collection. Three clinicians performed dual-adjudicated, retrospective chart review to identify the cause of respiratory failure in subjects enrolled between 2015–2020. Subjects were split into two cohorts: 1) 50 patients with confirmed moderate or severe ARDS diagnosed using Berlin consensus criteria within 7-days of intubation,[35] and 2) 50 patients with no evidence of ARDS during their course of MV (referred to here as non-ARDS).[20] Our cohort was split into clear phenotypes as a proof of concept study to test the feasibility of using VWD and ML to detect ARDS. Patients were excluded from the cohort if: 1) they were transferred to ICU from an outside facility; 2) had a chronic tracheostomy prior to admission; 3) had a persistent bronchopleural fistula; 4) diffuse chronic fibrotic lung disease; or 5) Berlin criteria were met for < 24 hours. Included patients had at least 1hr of VWD within either the first 24 hours of ventilation if they were non-ARDS, or 1 hour of VWD in the 24 hours after first meeting Berlin diagnostic criteria for ARDS. For specific cohort descriptive statistics see online supplement eTable 1.

Data Extraction

VWD representing air flow sampled at 50 Hz over time was extracted from the Puritan Bennet-840 mechanical ventilator using an established data collection architecture.[36] Our VWD sample contained 2,020,556 breaths, including 1,331,285 breaths from patients with ARDS and 689,271 breaths from patients without ARDS. For generation of model inputs, flow time series data were split into successive breath "instances", which we define as a sequence of flow samples that are 224 observations in length (Figure 1A). Breath instances were formatted so that each instance begins at the start of inhalation. We then arranged the data as a series of 20 breath instances, which we refer to as a "breath window", for our ARDS classification experiments.(Figure 1A) All breath windows were labeled according to their subject's corresponding pathophysiology (ARDS or non-ARDS).

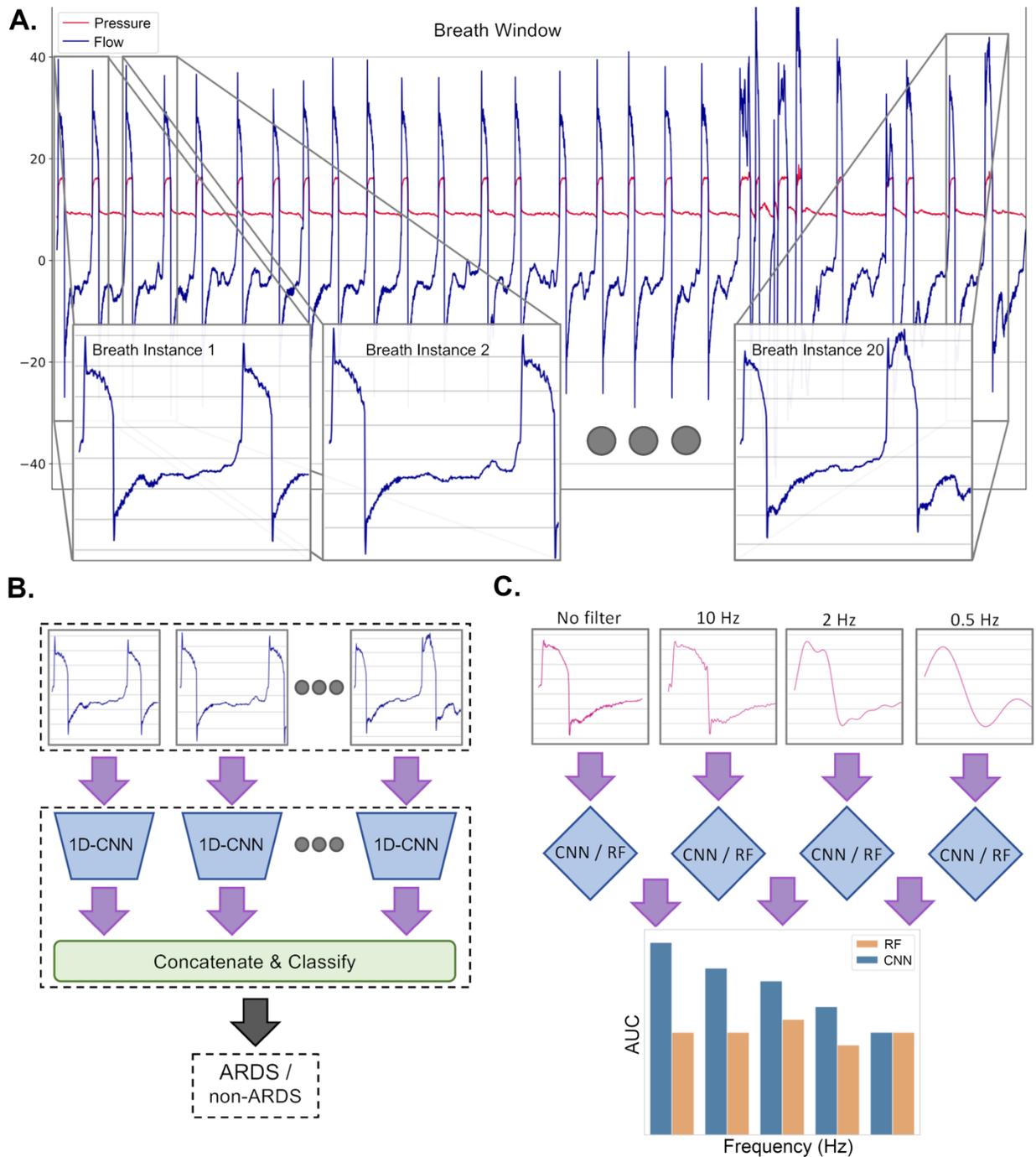

Figure 1: **A.** Ventilator data is extracted as a continuous series of raw flow data from the mechanical ventilator sampled at 50 Hz. **B.** Breath instances that were extracted from the above windows are fed into a 1-D convolutional neural network (CNN) to make an ARDS or non-ARDS classification on the instance. **C.** For interpretability, lowpass filtering was performed on the waveforms and then a CNN or random forest (RF) model was trained using the filtered data. CNN and RF model results were compared to determine how waveform data was utilized by the CNN.

Training & Visualization

Once we processed our data, we trained a one-dimensional (1-D), 18-layer, DenseNet CNN model to perform binary classification on which breath windows were classified as ARDS or non-ARDS.[37] ~~not.~~ Since there are > 1 breath windows per-patient, we can aggregate breath-window classifications to the patient by finding the majority classification of all breath windows for a patient. Thus, if > 50% of breath windows for a patient are classified as ARDS, then the patient was predicted as ARDS. Our primary reporting metric is model AUC based on the fraction of correctly-made patient-level predictions made for ARDS. We also report patient-level accuracy, sensitivity, and specificity.

We evaluated our model performance by first using the 5-fold K-fold splitting technique using data from 80 patients out of the 100-patient cohort, evenly split between ARDS and non-ARDS patients. Each fold is split evenly in terms of the ratio of non-ARDS patients to ARDS patients to avoid class imbalance at the patient level. We also evaluate a 70:30 holdout split, random K-fold splits, and bootstrapping with replacement at 80:20 split (see online supplement). Because the number of ARDS breath instances outnumbers non-ARDS by a 2:1 ratio, we correct this imbalance by performing random oversampling of non-ARDS breath instances.[38–40] Because of data noise, we used standard stochastic gradient descent instead of adaptive learning rate algorithms like ADAM.[41] We performed a grid search for optimal hyperparameters. Learning rate was found to be best at 0.001. To determine a consistent approximation of our model's performance we performed 10 trials of experiments for 10 epochs and then averaged the results to determine a final estimate of performance.[28]

We used multiple approaches for model interpretation. To understand the relative contributions of different components of the frequency domain to model performance, we transformed our VWD using the fast Fourier transform (FFT) from the SciPy library.[42] Since data was sampled at 50 Hz, FFT decomposes the signal into the frequency bandwidth between -25 and 25 Hz. FFT-transformed data were otherwise utilized in our model identically to unprocessed VWD in that breath instances were transformed into the frequency domain and then inputted into the model and classified. For model interpretability, we utilized Grad-Cam[43] to highlight a saliency map for VWD. We chose Grad-Cam in our application because it passed tests (i.e. data and model randomization tests) necessary for consideration as a model explanation method.[44] Grad-Cam was then used to visualize specific frequencies the model appeared to be using in classification.

Finally, we performed signal filtering experiments using FFT and threshold filtering to investigate the contributions of specific frequency components of the VWD signal on model performance. For these experiments we set a lowpass filter at a bandpass between 0.5-20Hz. VWD is then decomposed with FFT, and selected frequency ranges are removed. VWD is then reconstructed into the time-domain using an inverse FFT. Our model is then retrained on lowpass filtered data and standard model performance metrics are then reported. All code was developed using Python's scientific computing software suite and PyTorch.[45,46] All code is available (*https://github.com/hahnicity/deepards*) and anonymized datasets are available upon request.

# Results

We utilized 1-D flow time-series data to train our DL model. Because our deep neural network architecture requires a fixed number of input samples, we first explored the optimal input sequence length. Based on experimentation (see online supplement eTable 2 and 3), we found that breath instances of 224 inputs were optimal for input into our network. We then compared the performance of several DL architectures (CNN, LSTM, or hybrid CNN+LSTM) and found optimal performance with the CNN model (see online supplement eTable 4). We also found model performance improves when we ensure that each breath instance begins at the start of inhalation. We report results for our final model in eTable 6 across all 5 K-folds. Results from K-folds cross-validation were similar to results from experiments using a random 70:30 split of the dataset where 70% of patients' data were used for model development and 30% of patients were reserved for holdout testing (AUC 0.94±0.01), and to experiments using bootstrapping to split the dataset (AUC 0.91±0.012) (see online supplement eTable 5-7 and associated methods).

Table 1: Displays results from individual K-Folds for our convolutional neural network classifier. 95% CI are shown after the mean result of each metric from individual trials. AUC - area under the curve. PPV - positive predictive value (also known as precision). NPV - negative predictive value

| K-Fold Number | AUC | Accuracy | Sensitivity | Specificity | PPV | NPV |
|---|---|---|---|---|---|---|
| 1 | 0.99±0.01 | 0.85±0.056 | 1.0±0.0 | 0.7±0.113 | 0.78±0.07 | 1.0±0.0 |
| 2 | 0.88±0.025 | 0.79±0.069 | 0.65±0.15 | 0.92±0.037 | 0.9±0.06 | 0.74±0.094 |
| 3 | 0.94±0.018 | 0.85±0.037 | 0.9±0.075 | 0.8±0.113 | 0.83±0.068 | 0.9±0.067 |
| 4 | 0.98±0.005 | 0.89±0.019 | 0.85±0.05 | 0.92±0.037 | 0.92±0.042 | 0.86±0.037 |
| 5 | 0.97±0.013 | 0.85±0.031 | 1.0±0.0 | 0.7±0.075 | 0.77±0.045 | 1.0±0.0 |
| Mean of 5 k-folds | 0.95±0.019 | 0.84±0.026 | 0.88±0.068 | 0.81±0.06 | 0.84±0.038 | 0.9±0.05 |

Once we finalized our deep learning network, we compared the performance of our DL model to our previously published traditional ML model (Figure 2).[20]

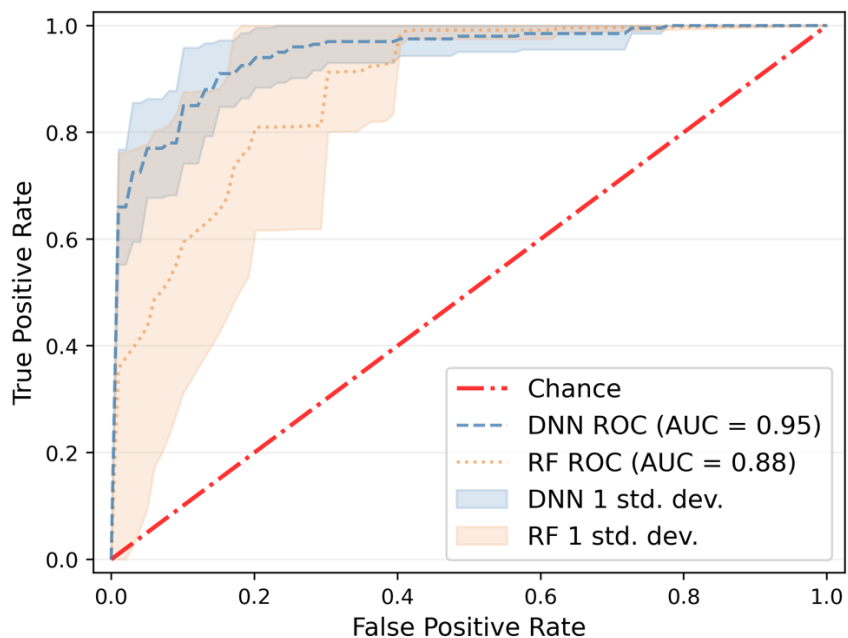

Figure 2: Comparison of ROC curves for both our initial random forest (RF) model, and our 1D-CNN model. Shaded parts of the plot are 95% CI interval.

After finding that our deep learning model outperformed the random forest model, we sought to investigate factors explaining the improved performance of deep learning. We began by exploring VWD components in the frequency domain that might be contributing to differential model performance by using 1-D FFT transformations of our data and then retraining the model with these transformed data (see online supplement eTable 8). When our network performed well with this experiment, we investigated the frequencies our network was using for classification. To do this, we utilized Grad-CAM to create a saliency map that highlights where in the frequency domain the network was focusing to make classifications, using all inputs into our neural network, and averaging all Grad-CAM outputs. The Grad-CAM saliency map shows that the model's cam intensities for ARDS were focused primarily on higher frequencies while non-ARDS classification was focused at both high and low-frequency bands.

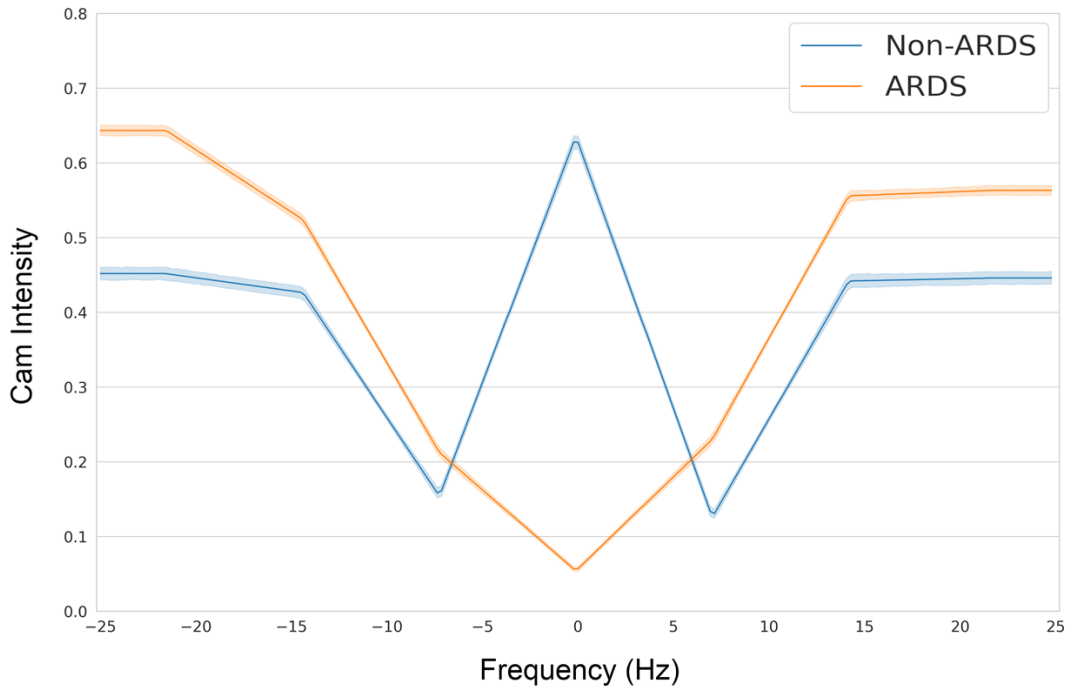

Figure 3: Shows an average Grad-CAM intensity (range 0-1.0) of our CNN that were trained only with FFT data. 95% CI is also shown in the slight boundary zone around each line. Non-ARDS focus is shown in blue, and ARDS in orange. Higher cam intensity values on the y-axis means that the network focused on this frequency more heavily for prediction of a specific class.

Based on the above observations, we sought to better understand the relative contributions of both high and low frequency components of the raw VWD signal to the improved performance observed in our DL model (**Error! Reference source not found.**). We thus performed targeted frequency ablation experiments by applying FFT to the raw VWD, with progressive lowpass filtering cutoff frequencies ranging from 0.5-20 Hz (see online supplement eFigure 1,2). The ablated VWD were then used for model development and K-folds cross validation using methods identical to those with the non-ablated raw VWD for both DL and traditional RF model development and cross-validation.[20] These frequency ablation experiments revealed a decrease in DL model performance after ablation of high frequency waveform components beginning at a 20 Hz lowpass cutoff down to a cutoff of 0.5 Hz, with performance comparable to our traditional ML approach at the cutoff of 0.5 Hz (**Error! Reference source not found.**B). Our RF model, which uses expert-derived features computed from raw VWD, did not see comparable decrements in performance at higher lowpass frequency cutoffs (**Error! Reference source not found.**C), which implies traditional ML features are well-represented in low frequency ranges. We were unable to ablate frequencies below 0.5Hz because we were unable to extract manual features for the traditional ML model because waveform ablation prevented the extraction of manual features (see online supplement eFigure 2).

Table 2: **Sub-table A** shows the baseline performances of the traditional machine learning based random forest (RF) model, and the convolutional neural network (CNN) model. **Sub-table B**. shows results of our network trained with only VWD after a lowpass FFT filter is applied. **Sub-table C** - Results of the traditional ML model after FFT lowpass filtering is applied. Note that AUC maintains similar performance through all ablations and matches performance of the FFT filtered model on CNN at very low frequencies.

| A. Baselines | Network | AUC | Accuracy | Sensitivity | Specificity |
|---|---|---|---|---|---|
| | RF[20] | 0.88±0.064 | 0.80±0.078 | 0.90±0.059 | 0.71±0.089 |
| | CNN | 0.95±0.019 | 0.84±0.026 | 0.88±0.068 | 0.81±0.06 |
| | | | | | |
| B. FFT Filtering on CNN | Cutoff Frequency (Hz) | AUC | Accuracy | Sensitivity | Specificity |
| | 20 | 0.94±0.015 | 0.83±0.017 | 0.85±0.038 | 0.81±0.035 |
| | 15 | 0.93±0.014 | 0.83±0.018 | 0.89±0.034 | 0.76±0.042 |
| | 10 | 0.92±0.021 | 0.81±0.026 | 0.84±0.039 | 0.77±0.046 |
| | 8 | 0.90±0.02 | 0.80±0.023 | 0.85±0.04 | 0.74±0.045 |
| | 6 | 0.91±0.021 | 0.81±0.026 | 0.85±0.037 | 0.78±0.042 |
| | 4 | 0.91±0.018 | 0.83±0.025 | 0.83±0.037 | 0.82±0.042 |
| | 2 | 0.91±0.016 | 0.82±0.025 | 0.81±0.037 | 0.83±0.04 |
| | 1 | 0.90±0.021 | 0.82±0.029 | 0.78±0.051 | 0.85±0.035 |
| | 0.5 | 0.89±0.015 | 0.81±0.18 | 0.77±0.03 | 0.86±0.031 |
| | | | | | |
| C. FFT Filtering on RF | Cutoff Frequency (Hz) | AUC | Accuracy | Sensitivity | Specificity |

|  | 20 | 0.88±0.063 | 0.83±0.074 | 0.86±0.068 | 0.79±0.08 |
|--|----|------------|------------|------------|-----------|
|  | 15 | 0.88±0.063 | 0.81±0.077 | 0.79±0.08 | 0.84±0.072 |
|  | 10 | 0.88±0.063 | 0.80±0.079 | 0.87±0.066 | 0.73±0.087 |
|  | 8  | 0.88±0.064 | 0.81±0.078 | 0.89±0.061 | 0.72±0.088 |
|  | 6  | 0.87±0.065 | 0.81±0.078 | 0.88±0.063 | 0.73±0.087 |
|  | 4  | 0.89±0.062 | 0.83±0.073 | 0.91±0.057 | 0.76±0.084 |
|  | 2  | 0.87±0.065 | 0.81±0.076 | 0.87±0.065 | 0.75±0.085 |
|  | 1  | 0.86±0.068 | 0.82±0.076 | 0.89±0.06 | 0.74±0.086 |
|  | 0.5 | 0.88±0.065 | 0.84±0.072 | 0.87±0.067 | 0.81±0.078 |

**Discussion**

In this study, we detail our work developing deep learning (DL) models for detecting ARDS using raw VWD as the only data source. We found that a 1-D deep CNN improved discrimination between ARDS and non-ARDS subjects compared to a random forest classical ML model, without need for labor-intensive and potentially biased expert-informed feature engineering. Furthermore, using a mix of descriptive and experimental methods to explain what our DL model was learning, we were surprised to find that relatively high-frequency regions of the 1-D VWD time series contained the implicit information that our DL model had learned to improve performance over our classical ML model.

Our work expands on previous research related to the development of algorithm-based screening for ARDS. Nearly every study to date has focused on processing data derived from the EHR including laboratory data, flowsheet data such as vital signs, text documents including, and radiographic images.[16] While early efforts used rule-based systems,[11–13,15,19] more recent studies have employed classical machine learning methods.[14,17,18] Many of these approaches have demonstrated good-excellent ability to discriminate between patients with and without ARDS at their originating centers, but several aspects of design potentially limit widespread applicability. In particular, reliance on infrequently sampled, non-continuous model inputs derived from clinical diagnostic testing may limit the timeliness of ARDS screening and/or model generalizability[10] as a function of clinician or institutional ordering and documentation practices. To address these potential limitations, recent work by our group used physiologic features extracted from VWD as inputs into a classical ML model to screen for ARDS.[20] Our results showed discrimination comparable to prior screening systems, even in the first six hours after Berlin diagnostic criteria for ARDS were first evident, without the need for additional model inputs from the EHR or other clinical systems. Despite the potential advantages of using VWD as the sole model input, our previous methods were limited by the need for expert-informed featurization of the raw VWD, which may have introduced biases into the model that compromised performance. Results of our current study suggest that deep learning may overcome limitations of our prior approach by learning from implicit information present in raw data whose value may not have been recognized by clinicians in the process of manual feature engineering.[20,27] While we are unaware of other studies directly comparing the performance of traditional ML and DL models operating on physiologic waveform data, multiple studies have demonstrated the potential utility of either classical ML or DL for classification and prediction using high sampling rate physiologic data.[47–49] Experimental comparisons of alternative ML approaches will be critical to developing a comparative effectiveness framework to guide future research while balancing competing considerations in terms of computational requirements and model development time, availability of technical and subject matter expertise, interpretability, regulatory complexity, and ease of deployment and maintenance.

One of the disadvantages of DL compared to some commonly used classical ML algorithms is a lack of interpretability, which can limit hypothesis-driven experimentation and compromise trust in model outputs.[27,50,51] Many works that have performed interpretability evaluations have utilized saliency methods.[52–55] This body of prior work motivated us to adopt Grad-CAM for our interpretability experiments. We also explored a number of different methods of ablating different parts of our VWD, or by searching for specific parts of the input that were salient, such as asynchronous breathing.[56] However, all these preliminary experiments failed to substantially impair the network's ability to learn. The failure of these analyses led to our investigation of using a lowpass FFT filter to remove high-frequency components from VWD.

Based on the results of these experiments, we saw that we were able to heavily filter VWD and still have it perform well in the task of ARDS detection. This suggests that there are key features that exist in the low-frequency representation of the waveform that are discriminative of ARDS.

To validate this question, we then performed learning on the lowpass filtered waveforms using featurization methods originally developed for traditional ML algorithms. This experiment builds upon prior work examining the importance of signal frequencies for predictive performance of machine learning models. In particular, there have been prior studies examining: the importance of spectroscopic wave spectrum for traditional machine learning;[57] the use of Grad-CAM to highlight important frequencies in sound recordings for bearing fault;[52] and Fourier transforms to improve interpretability of DNA models and atrial fibrillation detection.[58,59] We build upon this work by ablating model inputs over the frequency spectrum and then retrain our models with filtered data to obtain better understanding of how the models would perform. To the best of our knowledge however, no other surveyed studies have taken additional steps to retrain their models using filtered waveforms. We found that after retraining our traditional ML model on filtered waveforms model performance did not change. This shows us that the expert features that were originally derived exist on the low-frequency spectrum and are available to the DL model to learn. From observation of filtered waveforms (see online supplement eFigure 1,2), we note that low frequency data is associated with waveform periodicity, i.e. the rate at which a patient is breathing.  Given that DL model's performance is equivalent to performance of traditional ML when using heavily filtered waveforms, it is possible that our DL model is learning the same expert-derived features that were originally used for detecting ARDS. The fact that DL models are superior to traditional ML when higher frequency waveform data is included shows that DL models are likely able to learn discriminative features in higher-frequency data that would be difficult for experts to manually featurize. This observation fits broadly with the perceived benefit of using CNN,[28] and we hope our experiments will cause researchers to consider frequency ablation as a tool that may be used to better understand features that may or may not be present in PWD. Additional research into ventilator and other physiologic signals could also help to highlight useful features that are found in certain frequency bands.

Limitations

Our study has several limitations. First, our dataset was limited to a single academic medical center. Second, our input data was limited to a single physiologic data stream (flow), and it is unclear if our findings regarding the importance of high frequency signal components to model performance will generalize to other non-ventilator physiologic waveforms. Third, Grad-CAM provides low fidelity mappings for FFT inputs. This hinders our ability to reason about which specific frequencies are being heavily used in our network. SHAP or LIME may be better alternatives to Grad-CAM for this purpose, however currently, these tools are only available for use with 2-D image data for DL models and not 1-D time series. Our limited sample size and more restrictive cohort enrollment parameters may have resulted in model overfitting. We attempted to address this issue by using a 5 K-folds approach to model validation where we found relative stability of performance across K-folds, and with separate experiments using randomized K-folds. However, future research on larger, more diverse, patient cohorts will be required to determine the best uses of deep learning and VWD for ARDS screening. Finally, our subject selection criteria were chosen to ensure phenotypic separation between ARDS and non-ARDS subjects to test our hypothesis that VWD and ML could be used to discriminate between these clearly delineated phenotypes.[20] It is unclear how the performance of deep learning classifiers will be affected by the use of a more heterogeneous patient cohort, such as patients with unclear timing of ARDS onset, mild ARDS, or additional non-ARDS causes of respiratory failure.

## Conclusions

This study found deep learning models improve upon traditional ML models for detecting ARDS using raw VWD as the only data source. Furthermore, we performed frequency ablation study to highlight explicit frequencies of the ventilator waveform data the model was learning from. Our use of frequency ablation to highlight salient waveform information may have potential in future applications of physiologic waveform data so that scientists may be better able to understand the specific performance impact of different waveform frequencies for model development. Overall, our study provides additional rationale for incorporating ventilator waveform data into future ARDS detection models and gives evidence that deep learning may have advantages to expert-derived featurization in extracting physiologically useful information from raw ventilator waveform data.

# Supplement

<u>Descriptive Statistics</u>

eTable 3: Clinical characteristics of study subjects.

|  | ARDS (n=50) | Non-ARDS (n=50) |
|---|---|---|
| **Age** (median, IQR) | 57 (38-65) | 58 (49-67) |
| **Female** (%) | 13 (26) | 23 (46) |
| **BMI** (median, IQR) | 26.4 (22.3-33.8) | 25.9 (22.1-28.7) |
| **Obstructive lung disease** (n, %) | | |
| COPD | 0 (0) | 12 (24) |
| Asthma | 0 (0) | 5 (10) |
| **Reason for ICU admission** (n, %) | | |
| Acute hypoxemic respiratory failure | 24 (48) | - |
| COPD/asthma exacerbation | - | 17 (34) |
| Sepsis | 11 (22) | - |
| Metabolic encephalopathy/drug overdose | 2 (4) | 15 (30) |
| Airway edema/anaphylaxis | - | 5 (10) |
| Stroke | - | 4 (8) |
| Cardiac arrest | 9 (18) | 3 (6) |
| Heart Failure | - | 2 (4) |
| Upper gastrointestinal bleeding | - | 2 (4) |
| Trauma/Surgery | 3 (6) | 2 (4) |
| Pancreatitis | 1 (2) | - |
| **SOFA score** (median, IQR) | 13 (10-16) | 7.5 (5-10) |
| **Days from intubation to Berlin criteria** (median, IQR) | 0.1 (0.0-0.2) | NA |
| **Median PaO2/FiO2 first 24h** (median, IQR) | 176 (134-210) | 318 (267-423) |
| **Worst PaO2/FiO2 24h** (median, IQR) | 108 (66-137) | 278 (147-385) |
| **ARDS insult type** (n,%) | | |
| Pneumonia | 18 (36) | - |
| Aspiration | 14 (28) | - |
| Non-Pulmonary Sepsis | 10 (20) | - |
| Trauma | 2 (4) | - |
| Diffuse Alveolar Hemorrhage | 2 (4) | - |
| Pancreatitis | 1 (2) | - |
| Other | 3 (6) | - |
| **Hospital Length of Stay** (median, IQR) | 13.3 (6.6-25.4) | 7.0 (4.2-13-4) |
| **Hospital Mortality** (n, %) | 24 (48%) | 10 (20%) |
| **Ventilator- free days in 28 days** (median, IQR) | 6.6 (0-23.0) | 25.3 (10.6-26.9) |

Sequence Length

eTable 4 and eTable 5 summarize the cross-validation accuracy and AUC for different length sequences taken one at a time and given as an input to the neural network. After training, the neural network classifies each of these sequences in the test set as ARDS or non-ARDS, and then finds the majority voting (>0.5) to determine the final output class label for each patient. AUC is calculated in standard per-patient manner as explained in the main text. Scikit-learn provides function to evaluate the AUC.

eTable 4: Accuracy of model based on 5-fold cross validation.

| Fold N/length of sequence | 2048 | 1024 | 512 | 224 | 128 |
|---|---|---|---|---|---|
| Fold 1 | 0.8 | 0.8473 | 0.8210 | 0.8210 | 0.8315 |
| Fold 2 | 0.8052 | 0.8157 | 0.8 | 0.8157 | 0.7578 |
| Fold 3 | 0.8157 | 0.8052 | 0.7947 | 0.7894 | 0.8421 |
| Fold 4 | 0.8789 | 0.8789 | 0.8526 | 0.9315 | 0.9210 |
| Fold 5 | 0.7777 | 0.7666 | 0.7722 | 0.8222 | 0.8 |
| **Average** | *0.82* | *0.82* | *0.81* | *0.84* | *0.83* |

eTable 5: AUC of model based on 5-fold cross validation

| Fold N/length of sequence | 2048 | 1024 | 512 | 224 | 128 |
|---|---|---|---|---|---|
| Fold 1 | 0.8844 | 0.9522 | 0.9411 | 0.9522 | 0.9044 |
| Fold 2 | 0.9011 | 0.9277 | 0.9211 | 0.9377 | 0.8855 |
| Fold 3 | 0.8411 | 0.8566 | 0.8889 | 0.8805 | 0.9244 |
| Fold 4 | 0.9722 | 0.9577 | 0.94 | 0.9877 | 0.9266 |
| Fold 5 | 0.8156 | 0.79 | 0.8175 | 0.865 | 0.8937 |
| **Average** | *0.88* | *0.90* | *0.90* | *0.92* | *0.91* |

Network Type

Based on preliminary exploration to determine the best sequence length of VWD to input into CNN networks we found breath instances of 224 inputs were optimal for input into the CNN (eTable 4, eTable 5). For all experimentation using CNNs we use an 18-layer DenseNet network. Then, we explored whether CNN, LSTM, or a combined CNN+LSTM network was superior at ARDS detection. Our results (eTable 6) show that CNN performs slightly better than LSTM. When CNN is paired with LSTM, we saw LSTM did not contribute additional discriminative power to the network, possibly because there was no change in label across a breath window. Furthermore, CNN+LSTM combinations took longer to train and were more prone to overfitting compared to a standard CNN. Overfitting was also initially a problem with LSTM only, but we largely minimized this issue by utilizing variable length sequences instead of padding the end of our breaths.

eTable 6: We explore whether CNN, LSTM, or combined CNN+LSTM networks are best overall network architectures for performing inference on our dataset.

| Network | AUC | Accuracy | Sensitivity | Specificity |
|---|---|---|---|---|
| **CNN only** | **0.93** | **0.84** | 0.83 | **0.84** |
| LSTM only | 0.90 | 0.81 | 0.81 | 0.81 |
| CNN+LSTM | 0.92 | 0.79 | **0.95** | 0.62 |

Additional Data Splits

Besides using the 5-fold cross validation that we report in our manuscript, we also report results for different data splits. We show results for a 70/30 holdout data split that was explored in our prior publication. We show results for random K-fold splits. Random K-fold is performed in a manner where K-folds are split randomly for each trial and once all trials are complete results are averaged together. Finally, we performed a bootstrapped, 80/20 holdout split with replacement. The bootstrapping split is also done in similar methodology to our prior publication. It is important to note that our model performance decreased after the 3$^{rd}$ epoch of training in each of these circumstances. We report the results of the 10$^{th}$ epoch of training in accordance with our manuscript's methodology, and also at the epoch that the model performed best.

eTable 7: Holdout split results

| N Trials | Epoch | Accuracy | AUC | Sensitivity | Specificity |
|---|---|---|---|---|---|
| 10 trials | 10 | 0.83 | 0.91 | 0.91 | 0.75 |
| - | 3 (best) | 0.85 | 0.94 | 0.87 | 0.83 |

eTable 8: Random K-fold split results

| N Trials | Epoch | Accuracy | AUC | Sensitivity | Specificity |
|---|---|---|---|---|---|
| 50 trials | 10 | 0.84 | 0.93 | 0.86 | 0.82 |
| - | 2 (best) | 0.87 | 0.93 | 0.86 | 0.86 |

eTable 9: Bootstrapping, 80/20 split, with replacement

| N Trials | Epoch | Accuracy | AUC | Sensitivity | Specificity |
|---|---|---|---|---|---|
| 50 trials | 10 | 0.82 | 0.91 | 0.85 | 0.78 |
| - | 2 (best) | 0.85 | 0.91 | 0.86 | 0.84 |

Modeling of FFT data and evaluating GradCam outputs.

After showing that our deep learning model outperforms the traditional random forest model, we sought to investigate why it performs better. To explore this question, we performed a 1-D FFT transformation on each breath instance and retrained our neural network. Notably, by the 2$^{nd}$ epoch, our network performs surprisingly well with FFT. However, after the 2$^{nd}$ epoch, the network begins to overtrain. We then added the FFT signal to the VWD, and our network does not perform better than with VWD alone (eTable 10). This implies that the network is learning frequency-related information from the VWD itself and that the FFT data is not adding additional information to the system.

eTable 10: We show an examination of neural network performance when it is fed varying data types. We first note the traditional ML baseline that has already been established in the first row. In later rows we explore how varying input of VWD, and FFT transformed VWD causes our deep neural network performance to change. FFT data only approaches the performance of VWD-only network in epoch 2 before overtraining in later epochs. VWD & FFT data performs better than using FFT only, but still suffers from overtraining and does not improve upon the best deep NN model.

| Dataset | Deep NN? | Epoch | AUC | Accuracy | Sensitivity | Specificity |
| --- | --- | --- | --- | --- | --- | --- |
| VWD only (traditional ML) | ✘ | N/A | 0.88 | 0.80 | 0.90 | 0.71 |
| VWD only (best CNN) | ✓ | 10 | 0.95 | 0.84 | 0.88 | 0.82 |
| FFT only | ✓ | 2 | 0.90 | 0.82 | 0.85 | 0.78 |
| FFT only | ✓ | 10 | 0.87 | 0.76 | 0.79 | 0.72 |
| VWD & FFT | ✓ | 2 | 0.92 | 0.83 | 0.86 | 0.79 |
| VWD & FFT | ✓ | 10 | 0.89 | 0.80 | 0.84 | 0.75 |

Frequency Ablation

We show instances of separate breaths filtered in accordance to our FFT filter. The FFT filter operates by first performing an FFT transformation on VWD, then zeroing all coefficients outside the Hz passband. After this, the VWD is reconstituted using an inverse FFT transform. It is noted that the filter keeps many of the minute fluctuations inherent in the breaths even at 15Hz passband. We show this waveform for purposes of displaying the physiologic morphology of the breaths during the filtering. The filtered breaths maintain surprisingly detailed breath morphology even up to 6Hz. By around 2Hz we can see that breath characteristics begin to become smoothed. Inspiratory and expiratory morphology is still somewhat visible by 0.5Hz, however by 0.25Hz this morphology is almost completely removed. Because of this, we were unable to perform manual featurization of waveforms filtered with a 0.25 Hz lowpass filter.

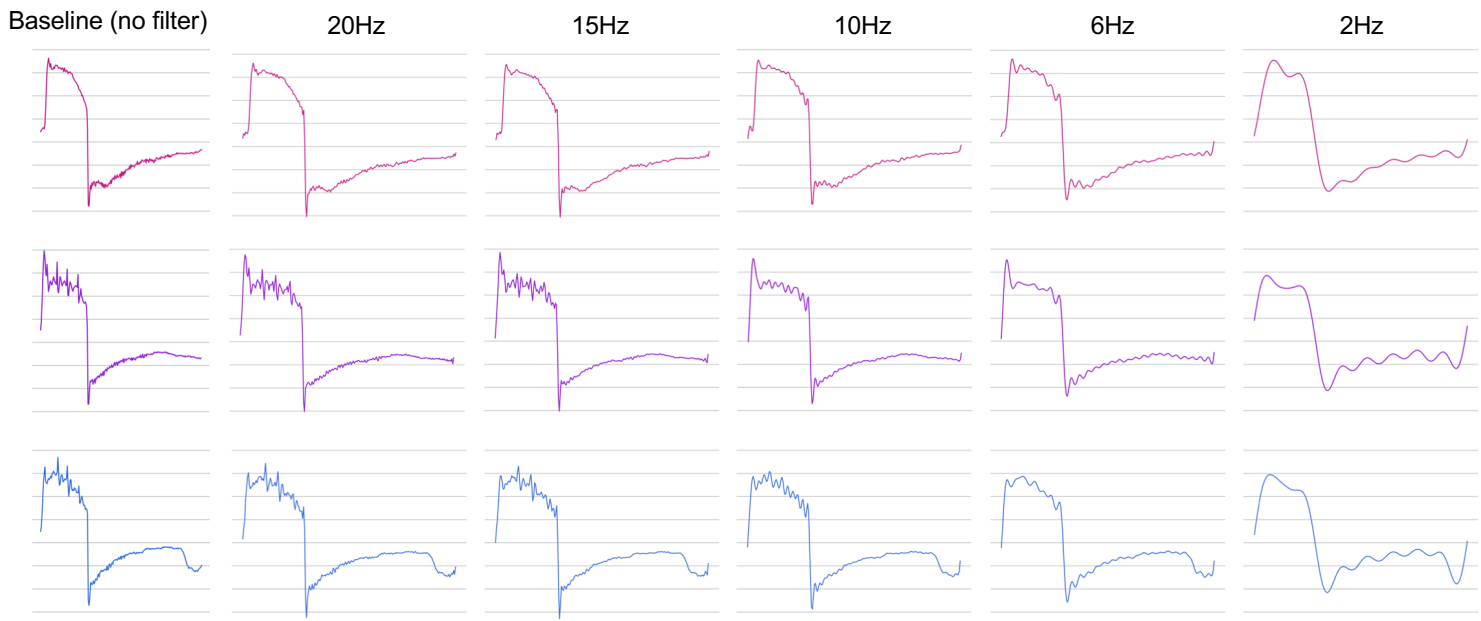

eFigure 1: We show examples of our 10th order lowpass FFT filter applied to 3 different breaths. We show only 20Hz, 15Hz, 10Hz, 6Hz, and 2Hz filters for content clarity.

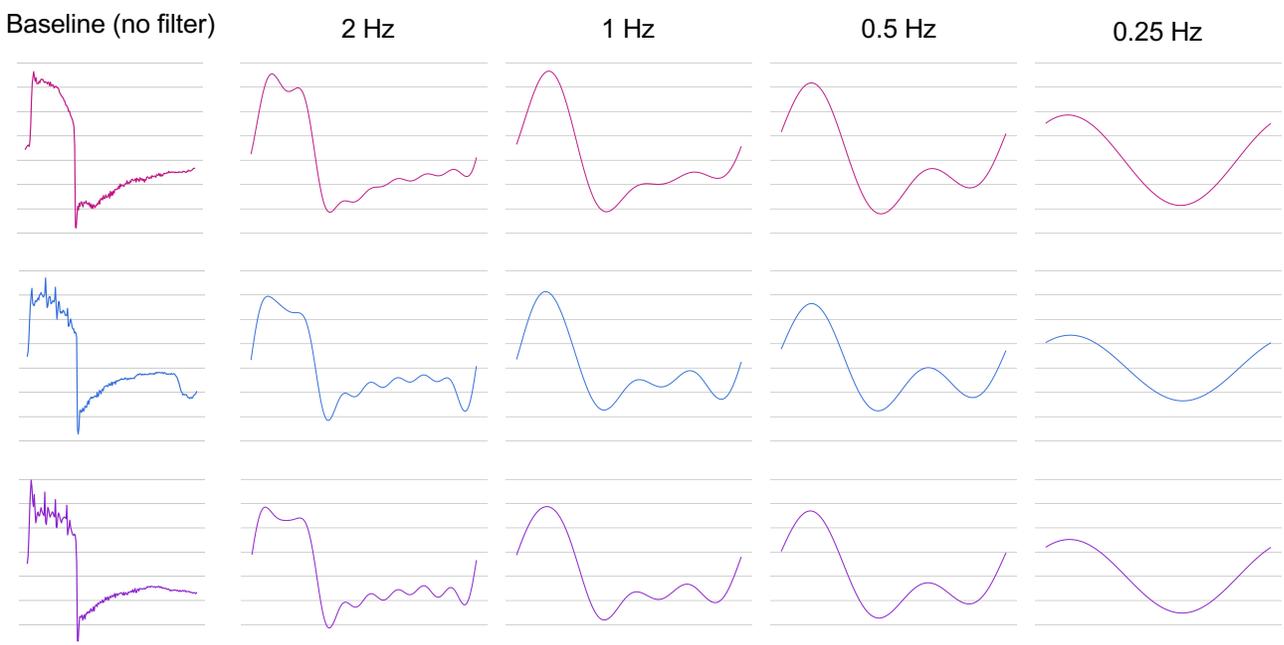

eFigure 2: We show examples of our lowpass FFT filter applied to 3 different breaths at lower frequencies. We show only 2 Hz, 1 Hz, 0.5 Hz, 0.25 Hz.